%% file: main.tex
\documentclass[letterpaper, 10 pt, conference]{ieeeconf}
\IEEEoverridecommandlockouts
\usepackage{amsmath,commath,amssymb,bbm,graphicx}

\usepackage{color,colortbl}
\usepackage[table]{xcolor}
\usepackage{threeparttable}

\usepackage{placeins}

\usepackage{multirow}
\usepackage{makecell}
\usepackage{booktabs}

\usepackage{float}

\usepackage{cite}
\usepackage{url}
\usepackage[hidelinks]{hyperref}
\hypersetup{
    colorlinks=true,
    linkcolor=black,
    filecolor=black,
    urlcolor=gray,
    citecolor=black,
    breaklinks=true
}

\usepackage{algorithm}
\usepackage[noend]{algpseudocode}
\algnewcommand{\LineComment}[1]{\State #1}

\definecolor{mylightgray}{gray}{0.9}
\definecolor{ds}{rgb}{0.0,0.8,0.1}
\newcommand{\na}{\cellcolor{mylightgray}}

\usepackage{tikz}
\usetikzlibrary{shapes,arrows}
\tikzstyle{decision} = [diamond, draw, fill=blue!20, 
    text width=4.5em, text badly centered, node distance=3cm, inner sep=0pt]
\tikzstyle{block} = [rectangle, draw,
    text width=5em, text centered, rounded corners, minimum height=4em]
\tikzstyle{line} = [draw, -latex']
\tikzstyle{cloud} = [draw, ellipse,fill=red!20, node distance=3cm,
    minimum height=2em]

\usepackage{pgfplots}
\pgfplotsset{width=10cm,compat=1.15}
\usepgfplotslibrary{statistics}
\usetikzlibrary{pgfplots.groupplots}

\newcommand*{\ie}{\textit{i.e.}}

\newcommand{\mtrx}[1]{\mathbf{#1}}

\usepackage[normalem]{ulem}

\definecolor{c1}{RGB}{0,119,187}  
\definecolor{c2}{RGB}{51,187,238} 
\definecolor{c3}{RGB}{0,153,136}  
\definecolor{c4}{RGB}{238,119,51} 
\definecolor{c5}{RGB}{204,51,17} 
\definecolor{c6}{RGB}{238,51,119} 
\definecolor{c7}{RGB}{187,187,187} 
\definecolor{c8}{RGB}{0 0 0} 

\title{\LARGE \bf
Towards safe human-to-robot handovers of unknown containers
}

\author{
    Yik Lung Pang,
    Alessio Xompero,
    Changjae Oh,
    Andrea Cavallaro
\thanks{Yik Lung Pang, Alessio Xompero, Changjae Oh, and Andrea Cavallaro are with Centre for Intelligent Sensing, Queen Mary University of London, UK \{{\tt\footnotesize y.l.pang}, {\tt\footnotesize a.xompero}, {\tt\footnotesize c.oh}, {\tt\footnotesize a.cavallaro}\}{\tt\footnotesize@qmul.ac.uk}.}
}%

\begin{document}

\maketitle
\thispagestyle{empty}
\pagestyle{empty}

\begin{abstract}
Safe human-to-robot handovers of unknown objects require accurate estimation of hand poses and object properties, such as shape, trajectory, and weight. Accurately estimating these properties requires the use of scanned 3D object models or expensive equipment, such as motion capture systems and markers, or both. However, testing handover algorithms with robots may be dangerous for the human and, when the object is an open container with liquids, for the robot. In this paper, we propose a real-to-simulation framework to develop safe human-to-robot handovers with estimations of the physical properties of unknown cups or drinking glasses and estimations of the human hands from videos of a human manipulating the container. We complete the handover in simulation, and we estimate a region that is not occluded by the hand of the human holding the container. We also quantify the safeness of the human and object in simulation. We validate the framework using public recordings of containers manipulated before a handover and show the safeness of the handover when using noisy estimates from a range of perceptual algorithms.
\end{abstract}

%
%

\section{Introduction}
\label{sec:intro}

Human-to-robot handovers are important for daily household activities. These interactions involve a human, a robot, and an object, and can be split into three phases, namely {human manoeuvring}, {handover}, and {robot manoeuvring}~\cite{Medina2016,Sanchez-Matilla2020}. For example, a human can give to the robot household containers, such as drinking glasses, cups, or food boxes, whose properties vary when filled with content. 
Therefore, an additional phase where the human prepares the object for the handover (object manipulation), e.g.~pouring content into a container, can be included prior to human manoeuvring. 
During human manoeuvring, the human holds the object and approaches the robot. The robot should understand the intention of the human to pass the object and approach the human hand. The robot should reach for a safe region on the object without harming the human, and close the gripper with sufficient forces to hold the object before the human releases their hold. During the handover, both the human and the robot are in contact with the object. During robot manoeuvring, the robot delivers the object to a target location without damaging the container or spilling the content.

Accurately estimating the locations of the human hands and the properties of the object is critical for the robot to safely execute handovers. Achieving this estimation with a real setup is challenging because no information about the object is available a priori to the robot, or additional and expensive equipment would be needed~\cite{Sanchez-Matilla2020}.
Referring specifically to household container-like objects, recent works classify the type and level of the content in a container or estimate the capacity and mass of the container (empty or filled), using audio or visual data~\cite{Donaher2021EUSIPCO_ACC,Ishikawa2020ICPR,Iashin2020ICPR,Liu2020ICPR,Modas2021ArXiv}. Reasoning about the human dynamics from visual data can also provide information about the physical properties of a container~\cite{duarte2020human}. 

Existing methods for human-to-robot handovers focus on the control algorithm of the robot to reach an object (e.g., water bottle) at a predicted location~\cite{Medina2016}. The safety of the human is partially addressed by avoiding to select grasp points near the human hand in an image~\cite{rosenberger2020object} or by classifying the human grasp type~\cite{Yang2020ArXiv_GraspClassification}. Moreover, most of the existing works rely on accurate predictions from expensive motion capture systems~\cite{Medina2016,nemlekar2019object}, or objects that are easily recognisable in the scene because of their shape or colour~\cite{vogt2018one}, or equipped with markers~\cite{pan2018evaluating}. Simulation environments provide a safe alternative to test robot algorithms for handovers. However, existing works focus on the planning of the handover trajectory and location~\cite{fishman2020collaborative,riccio2016learning,mainprice2012sharing}, disregard object shape and safety~\cite{Liu2020realsimreal}, and are limited to static setups where the robot waits for the object to be placed in the gripper~\cite{bestick2016implicitly,webster2020corroborative,vahrenkamp2016workspace}.

In this paper, we propose a framework that combines a perceptual algorithm using recordings of real scenes with a simulator to complete and achieve safe human-to-robot handovers. 
The framework addresses potential safety issues for the human and object before deploying the robot to a real environment, while requiring minimal hardware setup.
Our perceptual algorithm uses videos as input and estimates the poses of the human hands and the properties of a container, such as shape, weight, and location. Using these estimations, we propose to quantify the safeness of the handover as the probability for the robot gripper to touch the human hand (human safety) and the probability to drop, break or squeeze the container (object safety). Moreover, we estimate a safe grasp region that accounts for the available and unoccluded region on the held container to decide where to grasp without harming the human hand. We validate our framework on selected video recordings from the CORSMAL Containers Manipulation (CCM) dataset, where humans interact with different containers~\cite{Xompero_CCM}. Experiments show the safeness of the handovers when the physical properties are estimated with the vision algorithm or with different perceptual and multi-modal algorithms\footnote{Code, data, and videos of the experiments are available at: \url{http://corsmal.eecs.qmul.ac.uk/safe_handover.html}}.

\section{Related work}
\label{sec:relatedworl}

In this section, we review existing works using simulations for developing handover algorithms, addressing the safeness of the handover, and estimating the mass of containers.

Simulation platforms are often used by existing approaches for developing path planning algorithms focusing on the trajectory of the human hand and robot gripper, while providing a safe virtual environment before the deployment to a real setup~\cite{fishman2020collaborative,riccio2016learning,mainprice2012sharing}. However, the object and its properties are neglected~\cite{mainprice2012sharing}, precise 3D object models are required~\cite{bestick2016implicitly,webster2020corroborative,vahrenkamp2016workspace}, and grasping forces are not considered in the simulation~\cite{fishman2020collaborative,riccio2016learning}. 
Moreover, the handover is simulated in a static~\cite{webster2020corroborative} or a limited dynamic~\cite{bestick2016implicitly,vahrenkamp2016workspace} setup. In the former, the robot stays stationary while waiting for the object to be placed in the gripper. In the latter, the robot reaches a handover location, waits for the human to apply a force on the object, and then releases the gripper.
Alternatively, virtual and mixed reality combine user input from the real world and allow the users to interact with a virtual robot, but requiring additional hardware~\cite{fernandes2020human,dombrowski2017interactive,meyer2018improving}. A real-sim-real framework produces a task-relevant simulation environment based on visual estimations from RGB video, using simulated objects with primitive geometric shapes (spheres, cubes)~\cite{Liu2020realsimreal}. However, these approaches do not consider the human or object safety during a human-to-robot handover, which is crucial in a real-world scenario.

Safeness of the handover is important for the human and also for the object when it is fragile or deformable. Existing works have addressed the human safety by quantifying the risk of harming the human~\cite{kulic2007pre,rosenstrauch2017safe} or by designing strategies for the robot control to prevent collision between the human and robot~\cite{robla2017working,landi2019safety,zanchettin2015safety,Yang2020ArXiv_GraspClassification,rosenberger2020object}.
A danger index quantifies the safeness of the handover as the distance between the human and the robot, as well as the operating velocity and inertia of the robot~\cite{kulic2007pre}.
An exponentially decaying function is then modelled to reduce the danger index when the distance between the human and the robot is greater than a minimum threshold. This leads to a sharp decrease in the danger index when the robot is still close to the human.
The operating speed and force of the robot can be limited to reduce the risk of injuring humans in accidental collisions~\cite{rosenstrauch2017safe,landi2019safety}, as also outlined in ISO/TS 15066:2016~\cite{ISO15066}, or a safety barrier function can be defined around the robot~\cite{zanchettin2015safety}. 
A neural network is trained to classify the human hand pose into a discrete set of grasp types and then executes a corresponding predefined canonical robot grasp that is safe for the human~\cite{Yang2020ArXiv_GraspClassification}.
Potentially dangerous robot grasps that are too close to the human can be removed by segmenting the body and hand as well as applying a threshold on the pixel distance between the human and the grasp point~\cite{rosenberger2020object}. 
While the safety of the object has not yet been considered during the handover, there are existing works that assess the deformability of some objects. Deformability of the object can be minimised by defining an upper bound on the applied normal force using tactile data for each specific object~\cite{kaboli2016tactile}. A grasp metric that minimises the work done by the gripper jaws can be defined for deformable objects, but calibration against physical objects is still needed~\cite{xu2020minimal}.

Object mass estimation requires the reasoning on different physical properties, especially when the object is a container. Existing perception algorithms use uni-modal or multi-modal data, such as audio, images, and videos, to classify the content type and level as well as the container capacity~\cite{Iashin2020ICPR,Ishikawa2020ICPR,Liu2020ICPR,Modas2021ArXiv}. Convolutional neural networks can be trained to classify the content level within a range of containers from a single image when hand occlusions are present~\cite{Modas2021ArXiv}. While the performance is limited by the uni-modal input, the choice of training strategy, e.g.~combining adversarial training and transfer learning, can improve the classification accuracy~\cite{Modas2021ArXiv}. Independent classification of content type and level can be achieved by using convolutional and recurrent neural networks with only audio as input data~\cite{Ishikawa2020ICPR} or through late fusion of the predictions from both audio and visual features~\cite{Iashin2020ICPR}. Alternatively, multiple multi-layer perceptrons can be trained with audio data and conditioned on the container category estimated from a majority voting of the object detection across the frames of multi-view sequences~\cite{Liu2020ICPR}. 
Container capacity can be estimated as an approximation of a reconstructed shape~\cite{Ishikawa2020ICPR,Iashin2020ICPR,Xompero2020ICASSP_LoDE}. An iterative approach minimises a 3D primitive to the real object shape by constraining to the object segmentation mask from two views of a wide-baseline stereo camera, using both RGB, depth, and infrared images~\cite{Iashin2020ICPR}. The shape is then approximated to a cylinder to compute the capacity~\cite{Iashin2020ICPR}. 
The shape of a container can be instead reconstructed as a 3D point cloud by detecting the object in the RGB frame where the container is the most visible and using the corresponding pixel values in the depth map~\cite{Ishikawa2020ICPR}. Then, the shape can approximated to a cuboid and the capacity is computed by considering only the voxels that project within the object boundaries in the RGB image~\cite{Ishikawa2020ICPR}.
Alternatively, the prior knowledge of a detected object category can guide the sampling from a shape distribution using Gaussian process~\cite{Liu2020ICPR}. 
Despite achieving high accuracy in individual properties~\cite{Iashin2020ICPR,Ishikawa2020ICPR}, the estimation of the object mass is still challenging and a robot deployed in a real environment may apply forces that can be unsafe for the human or the object. 

\section{Problem definition}
\label{sec:problem_def}

Let $\mathcal{H}_L = \{ \mtrx{h}_{L,i}\}_{i=1}^{K}$ and $\mathcal{H}_R = \{ \mtrx{h}_{R,i}\}_{i=1}^{K}$, where $\mtrx{h}_{L,i},\mtrx{h}_{R,i} \in \mathbb{R}^3$, be the set of $K$ keypoints in 3D representing the left and right hand, respectively, of a human manipulating a container. The container is characterised by its dimensions given as maximum width $w$, height $h$, and depth $d$; its volume $V$; and its mass when empty $m_C$. 
During manipulation, the container is also represented by its time-varying location  $\mtrx{t}_k \in \mathbb{R}^3$ (i.e., the centroid in 3D) and orientation $\mtrx{\theta}_k \in \mathbb{R}^3$ at time $k$. The container can be either \textit{empty} or filled with a content $\tau \in \{\textit{pasta}, \textit{rice}, \textit{water}\}$ at different levels $\lambda \in \{\textit{half-full}, \textit{full}\}$, expressed as a percentage of the container capacity (i.e., 50\%, 90\%). This results in seven feasible combinations of content type and level. Moreover, each content type has an associated density $\rho(\tau)$.
The final object mass is defined as $m = m_C + m_f$, where  $m_f$ is the filling mass.

After manipulating the container, the human approaches the robot while holding the container to hand it over. The robot should approach the human and receive the container, while avoiding harming the human and stably holding the container. Once the handover is completed, the robot places the container at a pre-defined target area centred at $\mtrx{d} \in \mathbb{R}^3$ with radius $\eta$ on a table. 

The goal is to i) estimate the physical properties of a container, i.e.~width ($\tilde{w}$) and mass ($\tilde{m}$), ii) estimate the location of the container ($\tilde{\mathbf{t}}_k$) and human hands ($\tilde{\mathcal{H}}_{L,k}$, $\tilde{\mathcal{H}}_{R,k}$) over time, iii) perform a safe handover, and iv) quantify the safeness of the human and container.

\section{Safe human-to-robot handover}
\label{sec:perceptioncontrol}

To achieve a safe human-to-robot handover, we present a simulated handover framework recreated from the real scene based on the physical properties estimated by a perception algorithm\footnote{Note that we design the framework to be modular and efficient and hence different hardware and algorithms can replace the choices of our setup.}. We introduce an algorithm that identifies a target region on the container for the robot to grasp, while avoiding touching the human hand. We also quantify the safeness of the human and container during the handover within the simulation environment.

\begin{figure}[t!]
    \footnotesize
    \setlength\tabcolsep{2pt}
    \centering
    \begin{tabular}{cc}
        \includegraphics[trim={1.8cm 3.4cm 5cm 2.8cm},clip,width=1.0\columnwidth]{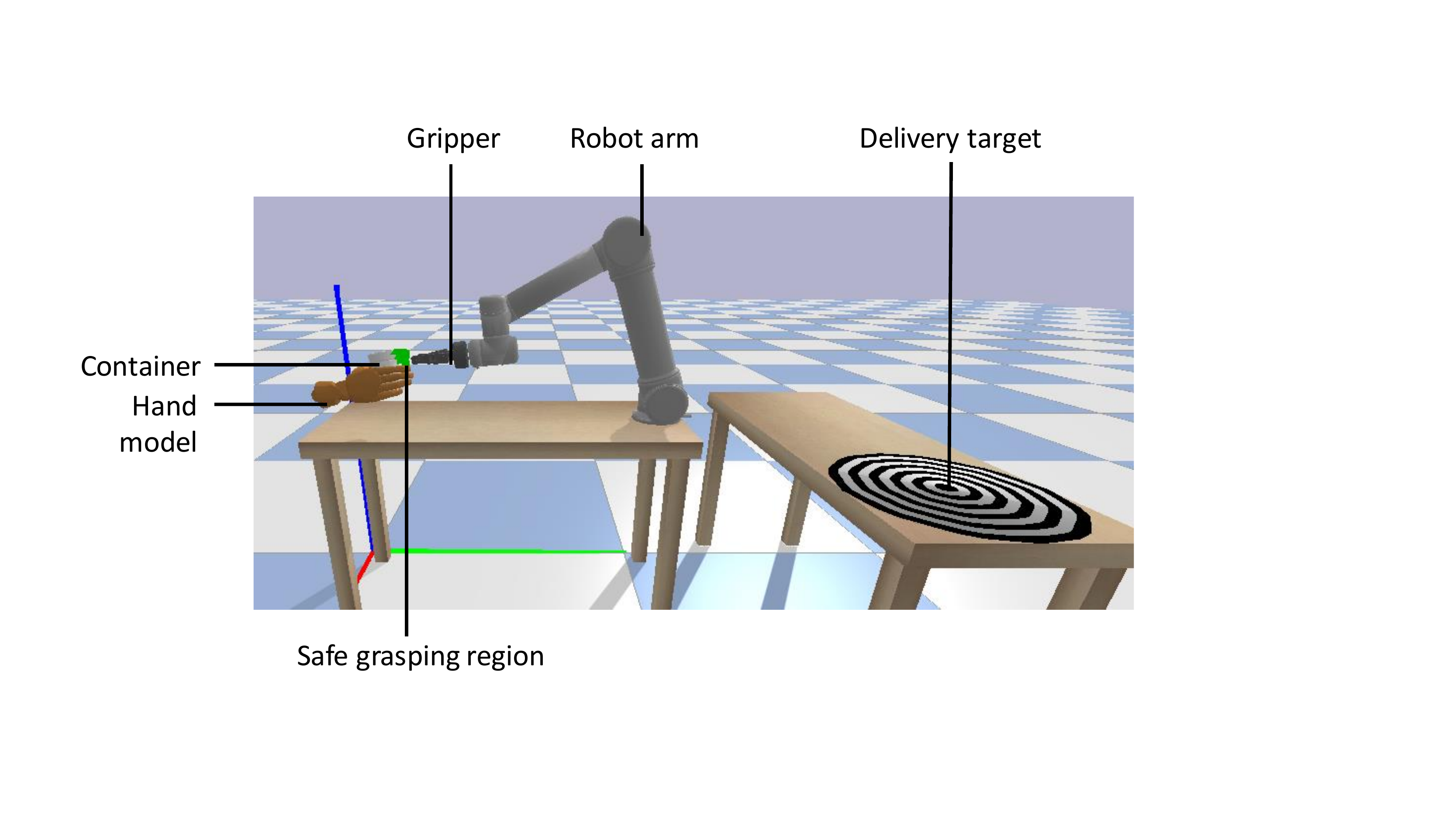}\\
    \end{tabular}
    \caption{The handover environment in PyBullet~\cite{coumans2019pybullet} and its coordinate system. The estimated safe grasp region on the container, highlighted in green, is not occluded by the hand and thus safe for the robot to receive the container without harming the human hand.
    }
    \label{fig:env_setup}
    \vspace{-5pt}
\end{figure}

\subsection{Handover setup}

We recreate in PyBullet~\cite{coumans2019pybullet} the handover setup that consists of the 6 DoF robotic arm (UR5) equipped with a 2-finger parallel gripper (Robotiq 2F-85); a table where object manipulation and  handover is happening, as well as where the robot is placed; an additional table behind the robot for delivering the object in a targeted area; a 3D container model; and left and right human hands based on the Modular Prosthetic Limb model~\cite{Johannes2011JHAPLTD_MPL}. Fig.~\ref{fig:env_setup} illustrates the handover setup in the simulator. 
The simulator renders the motion of the object and hands based on the predictions provided by a perception algorithm during object manipulation and human manoeuvring phases. The simulator also renders the motion of the robot arm based on a control algorithm, and then provides the applied forces while receiving the container in the moment of the simulated handover.

For the container, the simulator renders a 3D object model as reconstructed by the perception algorithm, assuming that scanned 3D models of the objects are not available.
To better replicate the real physical property, we use a high-precision electronic scale to annotate the object weight, with or without the content. 
We then simulate the container movement by defining a physical constraint that applies a fixed force to move the container to the estimated position and orientation.
Similarly for the hand motions, the simulator adjusts the hand and finger poses over time using a physical constraint based on the hand predictions, $\mathcal{H}_L$ and $\mathcal{H}_R$, and their estimated joint angles. 
The container movement is simulated until the robot reaches the container and closes the gripper to grasp the container. 
If the robot is not able to reach the container before the recording ends, we keep the last estimated location of both the container and the human hands for 2~s.

\subsection{Perception}

We estimate the pose of the hands over time, the location of a container over time, and the physical properties of the container, such as dimensions and mass, using only video inputs. Videos are acquired by fixed, calibrated stereo cameras, that capture object manipulation and human manoeuvring from the real scene. 

\begin{figure}[t!]
    \centering
    \setlength\tabcolsep{0pt}
    \begin{tabular}{ccc}
    \includegraphics[trim={3.5cm 1.8cm 4.5cm 2.8cm},clip,width=0.25\columnwidth,height=0.06\textheight]{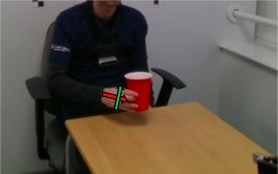} &
    \includegraphics[trim={2.2cm 1.2cm 3.2cm 2.0cm},clip,width=0.25\columnwidth,height=0.06\textheight]{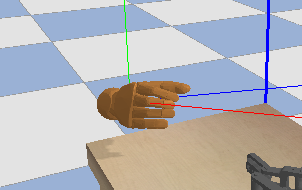}
    \end{tabular}
    \caption{Our manually defined {\protect\raisebox{2pt}{\protect\tikz \protect\draw[red,line width=2] (0,0.5) -- (0.3,0.5);}}~\textit{directional vector} and {\protect\raisebox{2pt}{\protect\tikz \protect\draw[green,line width=2] (0,0.5) -- (0.3,0.5);}}~\textit{up vector} (left). The {\protect\raisebox{2pt}{\protect\tikz \protect\draw[blue,line width=2] (0,0.5) -- (0.3,0.5);}}~\textit{right vector} is obtained as the vector orthogonal to both the directional and up vector. The reconstructed hand model (right).
    }
    \label{fig:hand_coords}
    \vspace{-10pt}
\end{figure}

To obtain the pose of the human hands, we first use OpenPose~\cite{simon2017hand} to detect the $K=21$ keypoints from each hand and for each camera view independently. For each keypoint, we triangulate its position in 3D~\cite{Hartley2003} and apply a Kalman filter~\cite{Kalman1960JBE} to reduce the impact of inaccurate estimations. We also approximate the hand model to a rigid object by defining directional vectors based on the estimated keypoints (see Fig.~\ref{fig:hand_coords}).

To obtain the trajectory of the container, we build upon an existing vision algorithm~\cite{Sanchez-Matilla2020} that localises and tracks a container from two camera views. We first localise the object of interest and initialise a 2D tracker in each view using Mask R-CNN~\cite{He2017ICCV_MaskRCNN}. Note that Mask R-CNN is pre-trained on COCO~\cite{Lin2018ECCV_COCO} considering only two classes, \textit{wine glass} and \textit{cup}. We then track the selected initial mask with SiamMask~\cite{Wang2019CVPR_SiamMask} in each view independently. For each frame $k$, we recover the 3D location ($\tilde{\mtrx{t}}_k$) of the container based on the triangulation method with the centroid of the two object masks (LoDE)~\cite{Xompero2020ICASSP_LoDE}. 
We then apply a Kalman filter~\cite{Kalman1960JBE} to $\tilde{\mtrx{t}}_k$ to reduce the impact of inaccurate estimations by LoDE, especially in the presence of occlusions. We assume that the hand and held container can be approximated to a single rigid object. Therefore, the orientation of the held container $\theta_k$ is set to the orientation of the human hand.

From the estimated 3D location, LoDE reconstructs a 3D shape of the container as a sparse point cloud,  \mbox{$\mathcal{Q} = \{ \mtrx{q}_j\}_{j=1}^J, \mtrx{q}_j \in \mathbb{R}^3$}, where $J$ is the total number of points. 
Assuming that containers have a rotationally symmetrical shape, LoDE approximates the shape to a set of 3D points lying on circumferences at different heights.
The radius of each circumference is obtained by minimising an initial hypothetical radius until all the 3D points of the circumference lie within the object mask of both views. From the estimated shape, the container dimensions $(\tilde{w},\tilde{h},\tilde{d})$ and volume $\tilde{V}$ can be obtained as a by-product, i.e.~the maximum width is given by the largest diameter among the circumferences, and the volume is approximated to the Riemman sum of the partial volumes (slicing method). 

To render in simulation a container as close as possible to the reality, we also perform offline shape estimation of the container with LoDE by manually selecting the frame where the container is occlusion-free.
We then process the sparse point cloud by defining faces of neighbouring points and forming a triangular 3D mesh. For some containers with concave shapes (e.g., wine and cocktail glasses), we use Volumetric Hierarchical Approximate Convex Decomposition~\cite{Mamou2016vhacd} to approximate the concave shape to multiple convex components. 

To estimate the object mass, we first determine the filling mass $\tilde{m}_{f,k}$ for the $k$-{th} frame as the product of estimated content level $\tilde{\lambda}_k$, volume $\tilde{V}$, and selected content density $\rho(\tilde{\tau}_k)$ based on the estimated type $\tilde{\tau}_k$:
\begin{equation}
    \tilde{m}_{f,k} = \tilde{\lambda}_k \tilde{V} \rho(\tilde{\tau}_k).
    \label{eq:fillingmass}
\end{equation}
Similar to~\cite{Modas2021ArXiv}, we devise a convolutional neural network-based classifier that predicts the content type and level for each frame and each view. 
In addition to the feasible combinations as classes, we also include \textit{unknown} as an extra class to handle opaque or translucent containers for which the content type and level cannot be estimated. Note that, prior to the hard decision on the predicted class, we multiply the predicted probabilities of all classes between the two views at the current frame $k$ and the predicted status at frame $k$ from the status at the previous frame $k-1$. For the latter, we design a fixed transition matrix that avoids the classification to change between unrelated statuses. Because of the object manipulation phase, we initialise equal probabilities for \textit{empty} and \textit{unknown} at $k=0$. Note that we only consider the estimation of the filling status at the handover moment.

We complement the estimations that cannot be obtained by our vision algorithm using the annotated values, such as the filling densities, $\rho(\tilde{\tau}_k)$, and container mass, $m_C$. 
When computing the filling mass, we use pre-defined filling densities as the average of the computed densities from the annotated masses, volumes, and filling levels: 0.41 g/mL for pasta, 0.85 g/mL for rice, and 1 g/mL for water. We then compute the estimated object mass as \mbox{$\tilde{m}_k = m_C + \tilde{m}_{f,k}$}.


\subsection{Robot control with safe grasp region}
\label{subsec:control}

\begin{figure}[t!]
    \centering
    \setlength\tabcolsep{0pt}
    \renewcommand{\arraystretch}{0}
    \begin{tabular}{ccc}
    \includegraphics[width=0.33\columnwidth,height=0.07\textheight]{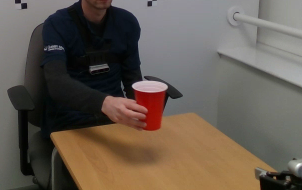} &
    \includegraphics[width=0.33\columnwidth,height=0.07\textheight]{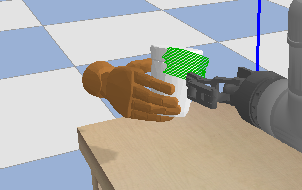} &
    \includegraphics[width=0.33\columnwidth,height=0.07\textheight]{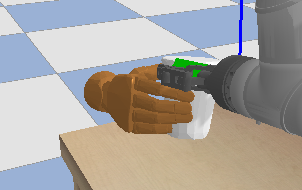} \\
    \includegraphics[width=0.33\columnwidth,height=0.07\textheight]{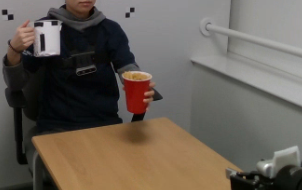} &
    \includegraphics[width=0.33\columnwidth,height=0.07\textheight]{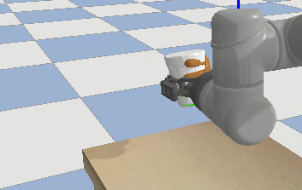} &
    \includegraphics[width=0.33\columnwidth,height=0.07\textheight]{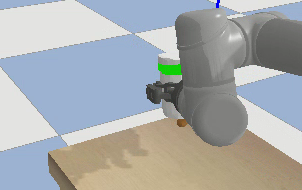} \\
    \includegraphics[width=0.33\columnwidth,height=0.07\textheight]{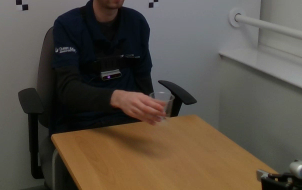} &
    \includegraphics[width=0.33\columnwidth,height=0.07\textheight]{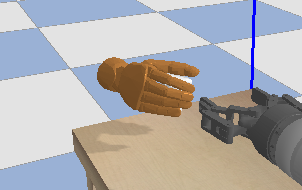} &
    \includegraphics[width=0.33\columnwidth,height=0.07\textheight]{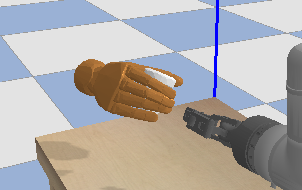} \\
    \end{tabular}
    \caption{Samples of safe regions and grasping. Top row: The estimated safe grasp region with successful grasping. Middle row: Target switch between disjoint safe regions, causing a collision with the hand as there is not enough time for the robot to react. Bottom row: The robot avoids approaching the container as there is no safe grasp region.}
    \label{fig:shaperendmpl}
    \vspace{-10pt}
\end{figure}

During the handover, the robot approaches the human to receive the object using the predicted time-varying object location $\tilde{\mtrx{t}}_k$ as input to the robot controller~\cite{Sanchez-Matilla2020}. However, $\tilde{\mtrx{t}}_k$ can be inaccurate and occluded by the human hand. We thus seek the region on the container surface that is not affected by the hand occlusion to execute the robot grasp. If a safe region is not available (see Fig.~\ref{fig:shaperendmpl}, bottom row), the control algorithm will keep the robot at a fixed distance from the container to guarantee the safety of the human.

To determine the safe region, we find a subset of the estimated object shape that is not occluded by the human hands and is in front of the robot. Moreover, we take into consideration a margin $r$ that accounts for both the width of the gripper $w_g$ and an enlarged thickness of the human finger $w_h$, i.e.~$r=(w_g + w_h)/2$. To this end, we first select a set of vertical components in $\mathcal{Q}$ that are within the margin of the gripper to grasp the object:
\begin{equation}
    \mathcal{Z} = \{ z : \min \mathcal{Q}_z + \frac{w_g}{2} \leq z \leq \max \mathcal{Q}_z - \frac{w_g}{2}\}.
\end{equation}
We then determine the set of unsafe vertical ranges based on the estimated keypoints of both human hands, \mbox{$\mathcal{H} \gets \mathcal{H}_{L} \cup \mathcal{H}_R$}, by selecting all the keypoints within the space occupied by the container:
\begin{equation}
   \Bar{\mathcal{Z}} = \{ h_z : (h_y > t_y - \frac{\tilde{w}}{2})  \land (\abs{h_x - t_x} < \tilde{w}) \},
\end{equation}
and we thus compute the safe range as
\begin{equation}
    \tilde{\mathcal{Z}} = \{ z \in \mathcal{Z} : (z \leq \min \Bar{\mathcal{Z}} - r) \lor (z \geq \max \Bar{\mathcal{Z}} + r)\}.
\end{equation}
The safe region is therefore given by all the 3D shape points in the safe range and in front of the robot, \mbox{$\{ \mtrx{q}_s \in \mathcal{Q} : q_{s,z} \in \tilde{\mathcal{Z}} \land q_{s,y} > t_y \}$}.
Note that there might be cases where the hand occludes the middle of the container resulting in two disjoint safe regions. In this situation, we select the largest subset of safe range because a small region can lead to higher chance of harming the human, e.g.~due to inaccuracies of the controller.

We integrate the safe region estimation into the control algorithm~\cite{Sanchez-Matilla2020} that continuously estimates and adjusts the joint angles of the robot arm to reach the container based on the prediction from the perceptual algorithm. Specifically, the control algorithm uses the Quadratic Programming solver with inverse kinematics formulations to reach the container at the predicted target location~\cite{Feng2015JFR}. To consider the estimated safe region, we modify the target location from the estimated 3D object centroid $\tilde{t}_k$ to the location $\mtrx{g} = [t_x, t_y, g_z]$ within the centre of the container and in the middle of the safe region, i.e.~$g_z =  (\max \tilde{\mathcal{Z}} + \min \tilde{\mathcal{Z}})/2$. Once the robot is within 1 cm from the target location, the robot closes the gripper with the estimated width from $\mathcal{Q}$ at $g_z$. The control algorithm will also target the orientation of the hand model as estimated by the vision algorithm to grasp the container directly from the front to avoid spilling of the content due to tilting of the gripper. 

To hold and deliver the container, the gripper must apply enough force to balance the weight of the object and the force introduced by the acceleration of the robot arm $a_{max}$ when holding and delivering the object. 
Therefore, we approximate the applied force to
\begin{equation}
    \tilde{F} \approx \frac{\tilde{m} (g + a_{max})}{\mu},
    \label{equ:grasp_force_theoretical}
\end{equation}
where $\tilde{m}$ is the predicted mass of the container (and content, if any); $g=9.81$~m/s$^{-2}$ is the gravitational acceleration; and $\mu$ is the coefficient of friction between the container and the gripper. To apply the required normal force $\tilde{F}$ given the predicted mass $\tilde{m}$, the robot needs to issue a joint effort $\tilde{\varepsilon}$. We design a linear model between an issued joint effort $\varepsilon$ and the normal force $F$ as applied by the robot in the simulator, 
\begin{equation}
    F = a\varepsilon + b.
    \label{eq:gripper_effort}
\end{equation}
We fit this model with a set of trials where a synthetic cylindrical object (length $10$~cm, radius $3$~cm, weight $0.01$~kg) placed in mid-air above a table was grasped and held by the robot arm to measure the applied total normal force, while varying the issued joint effort $\varepsilon$ ([$0.01, 1.25$]~N). The robot control issues $\tilde{\varepsilon}$ based on the predicted force $\tilde{F}$ and using the inverse model of Eq.~\ref{eq:gripper_effort}, where $\tilde{m}$ is estimated at the frame when the grasp is executed.

\input{safetymeasures}

\subsection{Safeness}

We assess the safeness of a handover within the simulation environment by separately quantifying two measures, human safety and object safety.

For \textit{human safety}, we quantify the probability of grasping a container while not touching the human hand. Specifically, we define a minimum safety distance $L$ around the human hand and a modified sigmoid function of $l$, \ie~the distance between the closest points of the robot and the human hand, 
\begin{equation}
    \psi_{h} = \frac{1}{1+e^{(\frac{2l}{L}-1) \ln{\frac{1-c}{c}}}},
    \label{eq:contactsafety}
\end{equation}
where $c$ controls the sensitivity of $\psi_h$ (Fig.~\ref{fig:grasp_contact}). We set the value of $c$ such that $\psi_h$ is converging at its maximum when the robot is further than the safety distance, i.e. $l\geq L$, and decays when too close to the human fingers, i.e.~$l < L$.

For \textit{object safety}, we quantify the probability that the predicted force applied by the robot will be able to hold the container so that it does not drop, break or deform. Specifically, we define an exponential function that accounts for the difference between the predicted normal force $\tilde{F}$ and the required normal force at the real object mass as available to the simulator, $\hat{F}$: 
\begin{equation}
    \psi_{f} = e^{\frac{\abs{\tilde{F}-\hat{F}}}{\hat{F}} \ln{(1-c)}},
    \label{eq:forcesafety}
\end{equation}
where $c$ controls the sensitivity of $\psi_f$ (Fig.~\ref{fig:grasp_contact}).
A negative difference represents an increase in the probability of dropping the container and a positive difference represents an increase in the probability of breaking or deforming the container. Note that the function penalises the safeness for rigid objects, as larger forces would be required to reach the breaking point compared to a non-rigid object that could start deforming with an applied force that is larger than the optimal one used to just hold the object~\cite{xu2020minimal,kaboli2016tactile}. 

\begin{figure}[t!]
    \centering
    \setlength{\tabcolsep}{0.1pt}
    \begin{tabular}{ccccccccc}
    \scriptsize{C1} & \scriptsize{C2} & \scriptsize{C3} & \scriptsize{C4} & \scriptsize{C5} & \scriptsize{C6} & \scriptsize{C7} & \scriptsize{C8} \\
    \includegraphics[width=0.12\columnwidth]{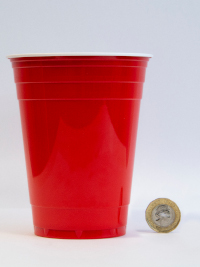}  &
    \includegraphics[width=0.12\columnwidth]{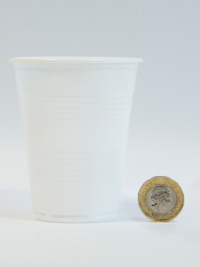} &
    \includegraphics[width=0.12\columnwidth]{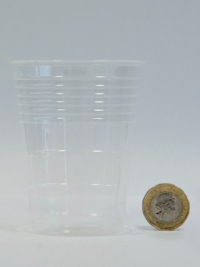} &
    \includegraphics[width=0.12\columnwidth]{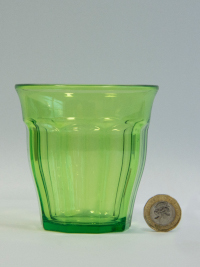} &
    \includegraphics[width=0.12\columnwidth]{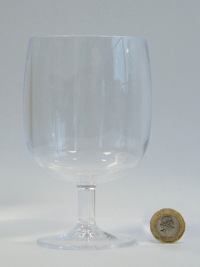} &
    \includegraphics[width=0.12\columnwidth]{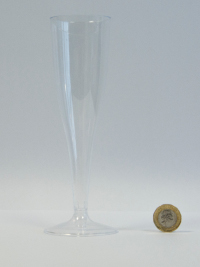} &
    \includegraphics[width=0.12\columnwidth]{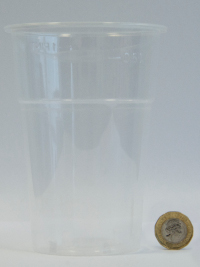} &
    \includegraphics[width=0.12\columnwidth]{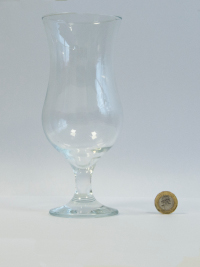} \\
    \includegraphics[width=0.12\columnwidth]{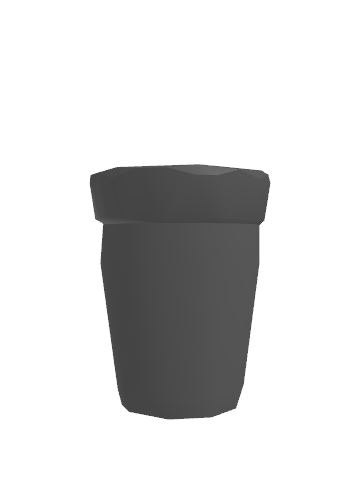} &
    \includegraphics[width=0.12\columnwidth]{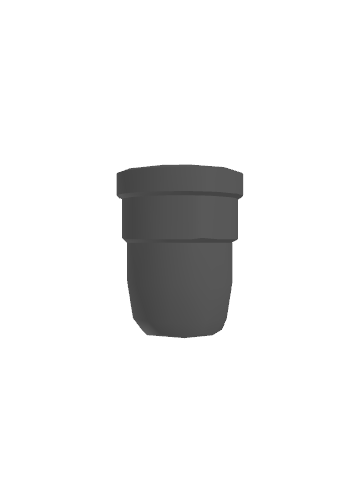} &
    \includegraphics[width=0.12\columnwidth]{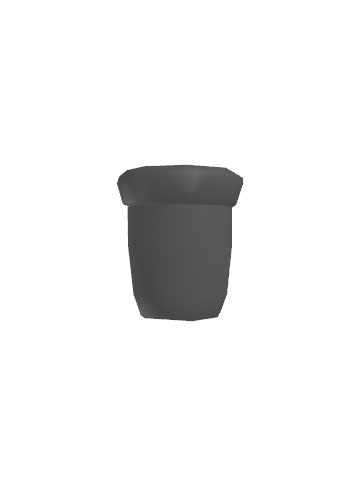} &
    \includegraphics[width=0.12\columnwidth]{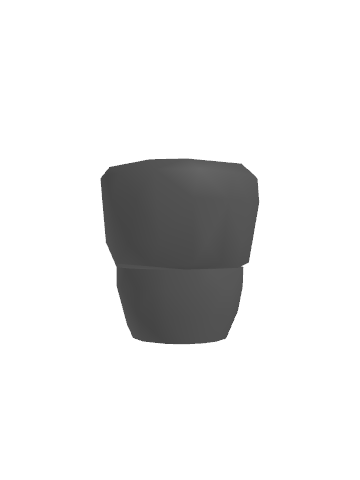} &
    \includegraphics[width=0.12\columnwidth]{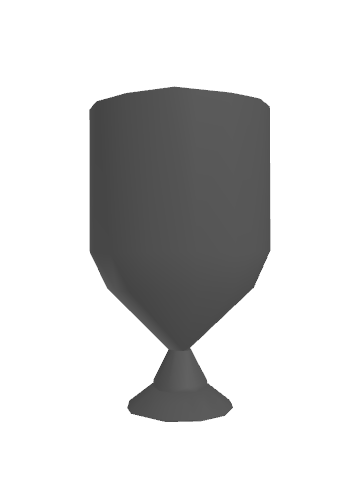} &
    \includegraphics[width=0.12\columnwidth]{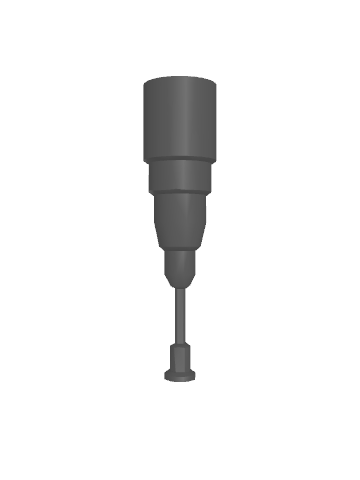} &
    \includegraphics[width=0.12\columnwidth]{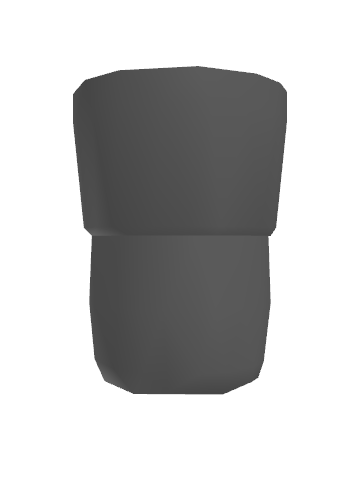} &
    \includegraphics[width=0.12\columnwidth]{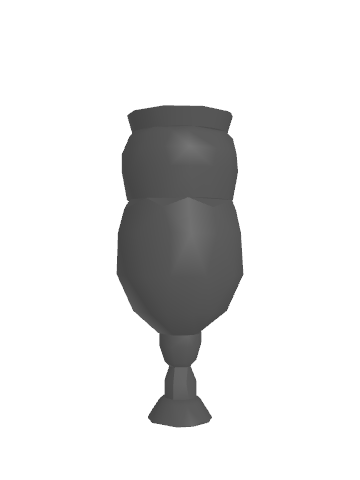} \\
    \end{tabular}
    \caption{Containers from the CCM dataset (top row) and their corresponding 3D mesh reconstruction (bottom row) used in our framework. Note the different container sizes compared to the \pounds1 coin used as reference size.
    }
    \label{fig:renderingredcup}
    \vspace{-10pt}
\end{figure}
\section{Validation}
\label{sec:validation}

\subsection{Experimental setup}

We evaluate the proposed framework on the CORSMAL Containers Manipulation dataset~\cite{Xompero_CCM}. We consider RGB videos, acquired at 30~Hz with 1280$\times$720 pixels resolution, from the two fully calibrated (intrinsic and extrinsic parameters) camera views fixed to the side of the robot arm. The videos contain a human sitting in front of the robot and pouring a content into a container placed on the table (scenario 1) or into a container while held by the human (scenario 2) before handing the container towards the robot. We select in total 112 recordings of eight cups and drinking glasses (see Fig.~\ref{fig:renderingredcup}, top row) captured under the same background and lighting conditions.

The recordings in the dataset contain only the object manipulation and human manoeuvring phases. The simulation environment of our framework complements the real recordings to complete the handover and robot manoeuvring phases, and assesses the safeness of the handovers. For further assessment of the handover safeness, we compared the physical properties estimated by our vision algorithm against those estimated by three alternative perceptual pipelines, namely BIT~\cite{Iashin2020ICPR},  HVRL~\cite{Ishikawa2020ICPR}, and VA2M~\cite{Liu2020ICPR} (see Sec.~\ref{sec:relatedworl} for details). Note that these alternative algorithms use both audio and visual data as input for the different tasks. Moreover, we use the estimations from our vision algorithm for the properties not addressed by these pipelines.

\subsection{Parameters setting}

We approximate the width of the human finger to $w_{h} = 2.0$~cm~\cite{dandekar20033} and set the width of the gripper finger to $w_{g} = 2.2$~cm ($r=2.1$~cm). We set $\mu = 1.0$ for both the hand and the container.
We set $L = r$ to discourage the gripper from colliding with the hand when reaching the container~\cite{kulic2007pre}.
We set $c = 0.995$ to ensure that the safety measures decay quickly as the probability of harm for the human and object increases. For $\Delta$, we set $\eta=500$~mm~\cite{Sanchez-Matilla2020}. We set $\phi=\frac{\pi}{4}$ as any container placed at this angle would fall after a few seconds before evaluating $\Delta$. For estimating the force $\tilde{F}$, we set $a_{max} = 27.9$ ms\textsuperscript{-2} (mean maximum acceleration). For the maximum joint effort, we obtain $a = 25.35$ and $b = 0.045$~N by fitting the model via linear regression.

\subsection{Performance measure}

We quantify the accuracy of delivering the container upright and within the target area as
\begin{equation}
    \Delta =
    \begin{cases} 
    1 - \frac{\alpha}{\eta}, & \text{if } (\alpha < \eta) \text{ and } (\beta < \phi), \\
    0, & \text{otherwise}, \\ 
    \end{cases}
\end{equation}
where $\alpha$ is the distance from the centre of the container base to the target location $\mtrx{d}$; $\eta$ is the maximum distance allowed from the pre-defined delivery location; $\beta$ is the angle between the vertical axis of the container and the vertical axis of the world coordinate system; and $\phi$ is the value of $\beta$ at which the container would tip over. 

\subsection{Results and discussion}
\label{subsec:results}

\input{sgp_safe_area}
In this section, we discuss the impact of selecting a safe grasp region on the safeness of the handover and compare different perception algorithms. The safety and performance measures were defined in the range [0,1] and will be discussed in this section as percentages.
Fig.~\ref{fig:sgp2} shows that estimating the safe region makes the robot control achieve safer handovers across the recordings and scenarios (points near the bottom-right corner). 
When the object centroid is not occluded by the hand, $\psi_h$ is close to 100 regardless of the use of the safe region (points near the top-right corner). When the visible area is too small ($\approx{w_g}$), grasping the container is challenging even when considering the safe region (points near the bottom-left corner). Note that there are a few cases where $\psi_h$ is low when selecting the safe region for grasping due to errors in the robot movements when reaching the target. Fig.~\ref{fig:sgp} shows that the algorithm achieves high human safety ($\psi_h>80$) across filling type and level configurations for wine glass (C5), champagne flute glass (C6), beer cup (C7), cocktail glass (C8). This is due to limited hand-occlusions, accurate object segmentation, or the control algorithm that keeps the robot at a safe distance when there are severe occlusions. Note that scenario 2 is more challenging than scenario 1 for both human and object safety measures due to hand occlusions at the start of the video leading to wrong segmentation of the object (failures shown by red crosses). 
Object safety is maximum for empty opaque containers, $\psi_f=100$, due to our assumption that the container is empty when we cannot estimate the content. Moreover, safeness increases for non-empty transparent containers, as the vision algorithm is more accurate when the filling is visible. Overall, our framework successfully grasps (high $\psi_f$) and accurately delivers ($\Delta>90$) the containers for 12 out of 112 recordings, while avoiding spilling (as we would expect in reality). Moreover, the robot does not perform approaching and container grasping for 35 out of 112 recordings due to low safeness and to ensure the safety of the human.

\input{table_test_containers}

In Tab.~\ref{tab:forcesafecomp1}, we compare the object safety based on the mass predicted by our vision algorithm against HVRL, BIT, and VA2M on the two CCM test sets (beer cup and cocktail glass) for scenario 1. Note that the object is occlusion-free in the first frames of the recordings of this scenario.
Correctly classifying filling type and level is challenging for our vision algorithm (ES) due to the varying container shapes and ambiguities, such as a container filled with water can be confused with an empty container. This results in a $\psi_f$ of around $30$ and an overestimation of $\tilde{m}$, indicating chances of deformation or breaking of the container in reality. 
By exploiting the audio modality, BIT, HVRL, and VA2M are often more accurate in estimating $\tilde{m}$, resulting in higher $\psi_f$. This is due to their ability to identify filling type and level by the characteristics and duration of the audio, even when the container is occluded by the hand. While object safety is maximum for empty containers \mbox{($\psi_f=100$)}, the probability decreases for cases with the presence of content. This is caused by the underestimation of the filling mass and hence we would expect the container to slip and fall, as the robot might not apply enough force. 
Moreover, we compute a complementary object safety $\bar{\psi}_f$ that replaces in Eq.~\ref{equ:grasp_force_theoretical} the predicted force with the applied total normal force measured in the simulator after grasp execution. 
This force includes additional factors that we did not model or control, e.g.~slipping between the contact surfaces and the shape of the container. Despite $\psi_f > 0$ before grasping execution, we can observe that there are cases where the robot fails to complete the execution of the grasp with the real mass (NE), e.g.~container filled with water up to 50\% due to the hand occluding the container. Moreover, there are many cases where object safety substantially differs before and after the grasping execution, e.g.~VA2M decreases from 95.22 to 0.25 or our vision algorithm increases from 13.16 to 27.10 for P5. This confirms that the object safety should consider the additional factors that may also occur in a real human-to-robot handover. Finally, we observe that the delivery accuracy is high ($\Delta \geq 80$) for all algorithms and all cases where the robot successfully held the container. This shows that the robot manoeuvring would have been stable enough to avoid spilling the content.

\begin{figure}[t!]
    \centering
    \setlength\tabcolsep{2pt}
    \begin{tabular}{cc}
    \includegraphics[trim={0 0 1cm 0},clip,width=0.595\columnwidth]{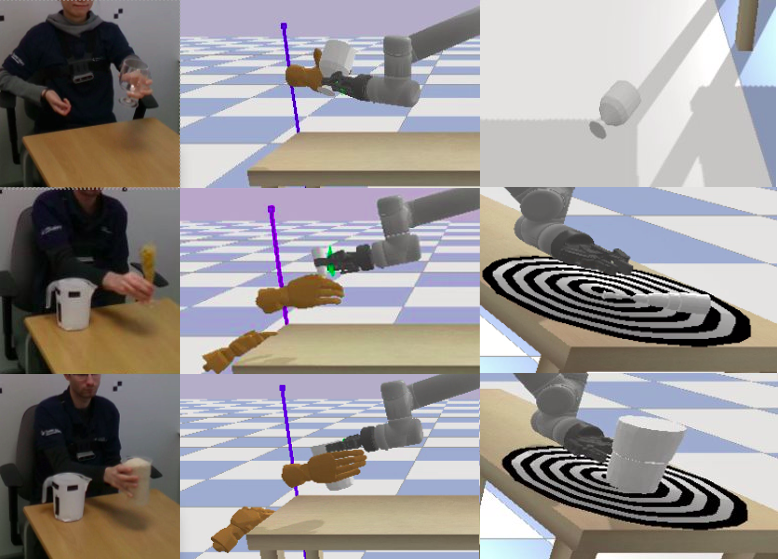} &
    \includegraphics[trim={0 0 1cm 0},clip,width=0.36\columnwidth]{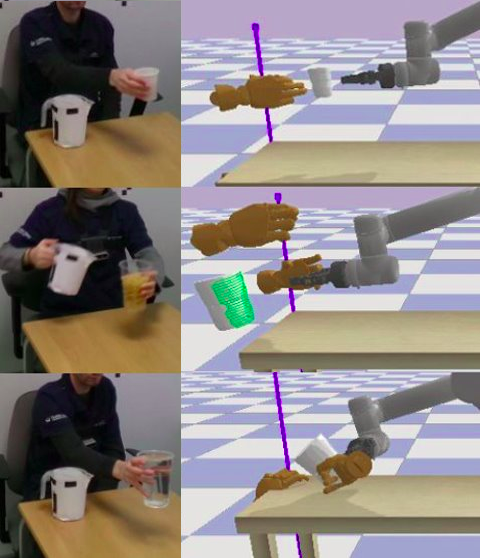} \\
    \end{tabular}
    \caption{Samples of human-to-robot handover results. Successful handovers (left) may end with unsuccessful delivery (dropped or tipped over), whereas unsuccessful handovers (right) are caused by failures of the vision algorithm (mismatch in hand and container position, incorrectly tracking of the jug instead of the container, inaccurate hand pose).}
    \label{fig:qualitativeres}
    \vspace{-10pt}
\end{figure}

In Fig.~\ref{fig:qualitativeres}, we show samples of human-to-robot handovers with our real-to-simulation frameworks, visualising the estimated safe region (in green) and the outcomes, such as successful receiving and holding, delivery, and failures.


\section{Conclusion}
\label{sec:conclusions}

We proposed a real-to-simulation framework that estimates the physical properties of unknown containers and their content from real videos of people manipulating the objects. The simulation part allows the framework to develop and assess algorithms for safe human-to-robot handovers of these objects. For human safety, we proposed to estimate the safe grasp region for the robot gripper to avoid the human hand when receiving the object. We also quantified the safeness of the human and object during the handover. 
We analysed the safety measures and tested our vision baseline compared to various approaches and sensing modalities (vision, sound).

Future work will include the validation of the algorithms discussed in this paper with a real setup as well as the simulation of deformable
objects and dynamics of the content.

\section*{ACKNOWLEDGMENT}

This work is supported by the CHIST-ERA program through the project CORSMAL, under UK EPSRC grant EP/S031715/1.

\bibliographystyle{IEEEtran}
\bibliography{IEEEabrv,main}



\end{document}

%% file: safetymeasures.tex
\pgfplotstableread{grasp_force_safety.txt}\graspforcesafety
\pgfplotstableread{grasp_contact_safety.txt}\sigmoid

\tikzstyle{every pin}=[
font=\tiny,]

\begin{figure}[t!]
    \centering
     \begin{tikzpicture}
        \begin{axis}[
            width=0.5\columnwidth,
            height=0.45\columnwidth,
            xlabel={$l$ (cm)},
            ylabel={$\psi_h$},
            ymin=0, ymax=1.1,
            xmin=0, xmax=5,
            xlabel near ticks,
            ylabel near ticks,
            label style={font=\footnotesize},
            tick label style={font=\scriptsize},
        ]
        \addplot+[solid, color=c1,mark=none] table[x=x_vals,y=y_vals1]{\sigmoid};
        \addplot+[solid, color=c2,mark=none] table[x=x_vals,y=y_vals2]{\sigmoid};
        \addplot+[solid, color=c4,mark=none] table[x=x_vals,y=y_vals3]{\sigmoid};
        \addplot+[solid, color=c5,mark=none] table[x=x_vals,y=y_vals4]{\sigmoid};
        \addplot+[dashed, color=gray,mark=none] coordinates{(2.1,0) (2.1,1.1)};
        \node[coordinate,pin=right:{$L=2.1$}] at (axis cs:2.1,0.3) {};
        \end{axis}
    \end{tikzpicture}
    \begin{tikzpicture}
        \begin{axis}[
            width=0.5\columnwidth,
            height=0.45\columnwidth,
            xlabel={$F (N)$},
            ylabel={$\psi_f$},
            ymin=0, ymax=1.1,
            xmin=0, xmax=20.0,
            xlabel near ticks,
            ylabel near ticks,
            label style={font=\footnotesize},
            tick label style={font=\scriptsize},
        ]
        \addplot+[solid, color=c7,mark=none] table[x=x_vals,y=y_vals1]{\graspforcesafety};
        \addplot+[solid, color=c1,mark=none] table[x=x_vals,y=y_vals2]{\graspforcesafety};
        \addplot+[solid, color=c2,mark=none] table[x=x_vals,y=y_vals3]{\graspforcesafety};
        \addplot+[solid, color=c4,mark=none] table[x=x_vals,y=y_vals4]{\graspforcesafety};
        \addplot+[solid, color=c5,mark=none] table[x=x_vals,y=y_vals5]{\graspforcesafety};
        \end{axis}
    \end{tikzpicture}
    \caption{Human and object safety functions. 
    Human safety, $\psi_h$, as a function of the distance between the closest points of the robot and the human subject's hand, $l$, when varying $c$: {\protect\raisebox{2pt}{\protect\tikz \protect\draw[c1,line width=2] (0,0.5) -- (0.3,0.5);}}~\textit{$c=0.5$};
    {\protect\raisebox{2pt}{\protect\tikz \protect\draw[c2,line width=2] (0,0.5) -- (0.3,0.5);}}~\textit{$c=0.7$};
    {\protect\raisebox{2pt}{\protect\tikz \protect\draw[c4,line width=2] (0,0.5) -- (0.3,0.5);}}~\textit{$c=0.9$};
    {\protect\raisebox{2pt}{\protect\tikz \protect\draw[c5,line width=2] (0,0.5) -- (0.3,0.5);}}~\textit{$c=0.995$}.
    Object safety, $\psi_f$, as a function of the total normal force applied on the object, $\tilde{F}$,  when $c = 0.995$ for plastic red cup (C1) with different content: 
    {\protect\raisebox{2pt}{\protect\tikz \protect\draw[c7,line width=2] (0,0.5) -- (0.3,0.5);}}~\textit{$Empty$};
    {\protect\raisebox{2pt}{\protect\tikz \protect\draw[c1,line width=2] (0,0.5) -- (0.3,0.5);}}~\textit{$50\%$~pasta};
    {\protect\raisebox{2pt}{\protect\tikz \protect\draw[c2,line width=2] (0,0.5) -- (0.3,0.5);}}~\textit{$90\%$~pasta};
    {\protect\raisebox{2pt}{\protect\tikz \protect\draw[c4,line width=2] (0,0.5) -- (0.3,0.5);}}~\textit{$50\%~$rice};
    {\protect\raisebox{2pt}{\protect\tikz \protect\draw[c5,line width=2] (0,0.5) -- (0.3,0.5);}}~\textit{$90\%$~pasta}.
    }
    \label{fig:grasp_contact}
    \vspace{-10pt}
\end{figure}

%% file: sgp_safe_area.tex
\pgfplotstableread{baseline_vs_sgp_s1.txt}\sgpcomparisona
\pgfplotstableread{baseline_vs_sgp_s2.txt}\sgpcomparisonb

\begin{figure}[t!]
    \centering
    \begin{tikzpicture}
        \begin{axis}[
            width=.5\columnwidth,
            xlabel={$\psi_h$ (SG)},
            ylabel={$\psi_h$ (No SG)},
            xmin=0, xmax=100,
            ymin=0, ymax=100,
            xlabel near ticks,
            ylabel near ticks,
            label style={font=\footnotesize},
            tick label style={font=\scriptsize},
        ]
        \addplot+[only marks, color=black, mark options={mark size=1.5pt,fill=c3}] table[x=SGP, y=Baseline]{\sgpcomparisona};
        \addplot[solid] coordinates {(0,0) (100,100)};
        \end{axis}
    \end{tikzpicture}
    \begin{tikzpicture}
        \begin{axis}[
            width=.5\columnwidth,
            xlabel={$\psi_h$ (SG)},
            ylabel={$\psi_h$ (No SG)},
            xmin=0, xmax=100,
            ymin=0, ymax=100,
            xlabel near ticks,
            ylabel near ticks,
            label style={font=\footnotesize},
            tick label style={font=\scriptsize},
        ]
        \addplot+[only marks, color=black, mark options={mark size=1.5pt,fill=c3}] table[x=SGP, y=Baseline]{\sgpcomparisonb};
        \addplot[solid] coordinates {(0,0) (100,100)};
        \end{axis}
    \end{tikzpicture}
    \caption{Comparison of human safety ($\psi_h$) between our safe grasping region estimation (SG) and the  baseline (No SG)~\cite{Sanchez-Matilla2020} on scenario 1 (left) and scenario 2 (right) from the CCM dataset. The more the points are near the bottom right, the more effective SG is. 
    }
    \label{fig:sgp2}
    \vspace{-10pt}
\end{figure}


\pgfplotstableread{heatmap_psi_h_s1.txt}\psihsa
\pgfplotstableread{heatmap_psi_h_s2.txt}\psihsb

\pgfplotstableread{heatmap_psi_f_s1.txt}\psifsa
\pgfplotstableread{heatmap_psi_f_s2.txt}\psifsb

\pgfplotstableread{heatmap_delta_s1.txt}\deltasa
\pgfplotstableread{heatmap_delta_s2.txt}\deltasb

\begin{figure}[t!]
    \centering
        \begin{tikzpicture}
        \begin{axis}[
            enlargelimits=false,
            axis on top,
            axis x line*=bottom,
            width=.5\columnwidth,
            view={0}{90},
            y dir=reverse,
            point meta min=0,
            point meta max=1,
            xmin=-0.5,xmax=7.5,
            ymin=-0.5,ymax=6.5,
            ytick={0,1,2,3,4,5,6},
            xtick={0,1,2,3,4,5,6,7},
            xticklabels=\empty,
            yticklabels={0,P5,P9,R5,R9,W5,W9},
            tick label style={font=\scriptsize},
            label style={font=\footnotesize},
            colormap={bw}{gray(0cm)=(1);gray(1cm)=(0);},
        ]
        \addplot [matrix plot*,mesh/cols=8, point meta=explicit] table [meta=z] {\psihsa};
        \addplot [only marks, mark=x, mark options={scale=2,draw=red}] coordinates {(7,3)  (7,4)};
        \end{axis}
        \begin{axis}[
            axis x line*=top,
            axis y line*=right,
            y dir=reverse,
            width=.5\columnwidth,
            view={0}{90},   
            enlargelimits=false,
            axis on top,
            point meta min=0,
            point meta max=1,
            xmin=-0.5,xmax=7.5,
            ymin=-0.5,ymax=6.5,
            ytick={0,1,2},
            xtick={0,1,2},
            xticklabels={},
            yticklabels={},
            tick label style={font=\scriptsize},
            yticklabel style = {},
            label style={font=\footnotesize},
         ]
         \end{axis}
    \end{tikzpicture}
        \begin{tikzpicture}
        \begin{axis}[
            enlargelimits=false,
            axis on top,
            axis x line*=bottom,
            width=.5\columnwidth,
            view={0}{90},
            y dir=reverse,
            colorbar,
            colorbar style={
                width=.02\columnwidth,
                ytick={0,0.2,0.4,0.6,0.8,1},
                yticklabels={0,20,40,60,80,100},
                ylabel={$\psi_h$},
            },
            point meta min=0,
            point meta max=1,
            xmin=-0.5,xmax=7.5,
            ymin=-0.5,ymax=6.5,
            ytick={0,1,2,3,4,5,6},
            xtick={0,1,2,3,4,5,6,7},
            xticklabels=\empty,
            yticklabels={},
            tick label style={font=\scriptsize},
            label style={font=\footnotesize},
            colormap={bw}{gray(0cm)=(1);gray(1cm)=(0);},
        ]
        \addplot [matrix plot*,mesh/cols=8, point meta=explicit] table [meta=z] {\psihsb};
        \addplot [only marks, mark=x, mark options={scale=2,draw=red}] coordinates {(5,0)  (4,1)  (5,1)  (2,3)  (5,3)  (2,4)  (7,4)  (5,5)  (7,5)  (2,6)  (5,6)};
        \end{axis}
        \begin{axis}[
            axis x line*=top,
            axis y line*=right,
            y dir=reverse,
            width=.5\columnwidth,
            view={0}{90},   
            enlargelimits=false,
            axis on top,
            point meta min=0,
            point meta max=1,
            xmin=-0.5,xmax=7.5,
            ymin=-0.5,ymax=6.5,
            ytick={0,1,2},
            xtick={0,1,2},
            xticklabels={},
            yticklabels={},
            tick label style={font=\scriptsize},
            yticklabel style = {},
            label style={font=\footnotesize},
         ]
         \end{axis}
    \end{tikzpicture}
    %
        \begin{tikzpicture}
        \begin{axis}[
            enlargelimits=false,
            axis on top,
            axis x line*=bottom,
            width=.5\columnwidth,
            view={0}{90},
            y dir=reverse,
            point meta min=0,
            point meta max=1,
            xmin=-0.5,xmax=7.5,
            ymin=-0.5,ymax=6.5,
            ytick={0,1,2,3,4,5,6},
            xtick={0,1,2,3,4,5,6,7},
            xticklabels=\empty,
            yticklabels={0,P5,P9,R5,R9,W5,W9},
            tick label style={font=\scriptsize},
            label style={font=\footnotesize},
            colormap={bw}{gray(0cm)=(1);gray(1cm)=(0);},
        ]
        \addplot [matrix plot*,mesh/cols=8, point meta=explicit] table [meta=z] {\psifsa};
        \addplot [only marks, mark=x, mark options={scale=2,draw=red}] coordinates {(7,3)  (7,4)};
        \end{axis}
        \begin{axis}[
            axis x line*=top,
            axis y line*=right,
            y dir=reverse,
            width=.5\columnwidth,
            view={0}{90},   
            enlargelimits=false,
            axis on top,
            point meta min=0,
            point meta max=1,
            xmin=-0.5,xmax=7.5,
            ymin=-0.5,ymax=6.5,
            ytick={0,1,2},
            xtick={0,1,2},
            xticklabels={},
            yticklabels={},
            tick label style={font=\scriptsize},
            yticklabel style = {},
            label style={font=\footnotesize},
         ]
         \end{axis}
    \end{tikzpicture}
        \begin{tikzpicture}
        \begin{axis}[
            enlargelimits=false,
            axis on top,
            axis x line*=bottom,
            width=.5\columnwidth,
            view={0}{90},
            y dir=reverse,
            colorbar,
            colorbar style={
                width=.02\columnwidth,
                ytick={0,0.2,0.4,0.6,0.8,1},
                yticklabels={0,20,40,60,80,100},
                ylabel={$\psi_f$},
            },
            point meta min=0,
            point meta max=1,
            xmin=-0.5,xmax=7.5,
            ymin=-0.5,ymax=6.5,
            ytick={0,1,2,3,4,5,6},
            xtick={0,1,2,3,4,5,6,7},
            xticklabels=\empty,
            yticklabels={},
            tick label style={font=\scriptsize},
            label style={font=\footnotesize},
            colormap={bw}{gray(0cm)=(1);gray(1cm)=(0);},
        ]
        \addplot [matrix plot*,mesh/cols=8, point meta=explicit] table [meta=z] {\psifsb};
        \addplot [only marks, mark=x, mark options={scale=2,draw=red}] coordinates {(5,0)  (4,1)  (5,1)  (2,3)  (5,3)  (2,4)  (7,4)  (5,5)  (7,5)  (2,6)  (5,6)};
        \end{axis}
        \begin{axis}[
            axis x line*=top,
            axis y line*=right,
            y dir=reverse,
            width=.5\columnwidth,
            view={0}{90},   
            enlargelimits=false,
            axis on top,
            point meta min=0,
            point meta max=1,
            xmin=-0.5,xmax=7.5,
            ymin=-0.5,ymax=6.5,
            ytick={0,1,2},
            xtick={0,1,2},
            xticklabels={},
            yticklabels={},
            tick label style={font=\scriptsize},
            yticklabel style = {},
            label style={font=\footnotesize},
         ]
         \end{axis}
    \end{tikzpicture}
    %
        \begin{tikzpicture}
        \begin{axis}[
            enlargelimits=false,
            axis on top,
            axis x line*=bottom,
            width=.5\columnwidth,
            view={0}{90},
            y dir=reverse,
            point meta min=0,
            point meta max=1,
            xmin=-0.5,xmax=7.5,
            ymin=-0.5,ymax=6.5,
            ytick={0,1,2,3,4,5,6},
            xtick={0,1,2,3,4,5,6,7},
            xticklabels={C1,C2,C3,C4,C5,C6,C7,C8},
            yticklabels={0,P5,P9,R5,R9,W5,W9},
            tick label style={font=\scriptsize},
            label style={font=\footnotesize},
            colormap={bw}{gray(0cm)=(1);gray(1cm)=(0);},
        ]
        \addplot [matrix plot*,mesh/cols=8, point meta=explicit] table [meta=z] {\deltasa};
        \addplot [only marks, mark=x, mark options={scale=2,draw=red}] coordinates {(7,3)  (7,4)};
        \end{axis}
        \begin{axis}[
            axis x line*=top,
            axis y line*=right,
            y dir=reverse,
            width=.5\columnwidth,
            view={0}{90},   
            enlargelimits=false,
            axis on top,
            point meta min=0,
            point meta max=1,
            xmin=-0.5,xmax=7.5,
            ymin=-0.5,ymax=6.5,
            ytick={0,1,2},
            xtick={0,1,2},
            xticklabels={},
            yticklabels={},
            tick label style={font=\scriptsize},
            yticklabel style = {},
            label style={font=\footnotesize},
         ]
         \end{axis}
    \end{tikzpicture}
        \begin{tikzpicture}
        \begin{axis}[
            enlargelimits=false,
            axis on top,
            axis x line*=bottom,
            width=.5\columnwidth,
            view={0}{90},
            y dir=reverse,
            colorbar,
            colorbar style={
                width=.02\columnwidth,
                ytick={0,0.2,0.4,0.6,0.8,1},
                yticklabels={0,20,40,60,80,100},
                ylabel={$\Delta$},
            },
            point meta min=0,
            point meta max=1,
            xmin=-0.5,xmax=7.5,
            ymin=-0.5,ymax=6.5,
            ytick={0,1,2,3,4,5,6},
            xtick={0,1,2,3,4,5,6,7},
            xticklabels={C1,C2,C3,C4,C5,C6,C7,C8},
            yticklabels={},
            tick label style={font=\scriptsize},
            label style={font=\footnotesize},
            colormap={bw}{gray(0cm)=(1);gray(1cm)=(0);},
        ]
        \addplot [matrix plot*,mesh/cols=8, point meta=explicit] table [meta=z] {\deltasb};
        \addplot [only marks, mark=x, mark options={scale=2,draw=red}] coordinates {(5,0)  (4,1)  (5,1)  (2,3)  (5,3)  (2,4)  (7,4)  (5,5)  (7,5)  (2,6)  (5,6)};
        \end{axis}
        \begin{axis}[
            axis x line*=top,
            axis y line*=right,
            y dir=reverse,
            width=.5\columnwidth,
            view={0}{90},   
            enlargelimits=false,
            axis on top,
            point meta min=0,
            point meta max=1,
            xmin=-0.5,xmax=7.5,
            ymin=-0.5,ymax=6.5,
            ytick={0,1,2},
            xtick={0,1,2},
            xticklabels={},
            yticklabels={},
            tick label style={font=\scriptsize},
            yticklabel style = {},
            label style={font=\footnotesize},
         ]
         \end{axis}
    \end{tikzpicture}
    \caption{Safety measures for human ($\psi_h$) and object ($\psi_f$, $\Delta$) across the 112 CCM recordings. Note that
    failures in detecting or estimating the hand pose are shown with a red cross. KEY -- 0:~empty; FX:~filling type (F) and level (X), where F represents pasta (P), rice (R), or water (W); and X is  half-full (5) or full (9). CY:~container (C) id (Y) in the same order as Fig.~\ref{fig:renderingredcup}.
    }
    \label{fig:sgp}
    \vspace{-10pt}
\end{figure}


%% file: table_test_containers.tex
\begin{table}[t!]
    \setlength\tabcolsep{0.8pt}
    \footnotesize
    \centering
    \caption{Comparison of object safety before grasp execution ($\psi_f$) and after grasp execution ($\bar{\psi}_f$), delivery accuracy ($\Delta$), and bias in the estimated object mass ($\delta_m$) between our vision algorithm (ES) and alternative perception algorithms on CCM test sets (scenario 1). 
    Failure cases are shown in gray. key~-~ne:~no grasp executed on the real mass.}
    \begin{tabular}{cc rrrr rrrr}
        &  &
        \multicolumn{4}{c}{\includegraphics[width=0.12\columnwidth]{beer_cup.jpg}} &
        \multicolumn{4}{c}{\includegraphics[width=0.12\columnwidth]{cocktail_glass.jpg}} \\
        \specialrule{1.2pt}{0.2pt}{1pt}
        & & \multicolumn{1}{r}{ES} & \multicolumn{1}{r}{BIT} & \multicolumn{1}{r}{HVRL} & \multicolumn{1}{r}{VA2M} & \multicolumn{1}{r}{ES} & \multicolumn{1}{r}{BIT} & \multicolumn{1}{r}{HVRL} & \multicolumn{1}{r}{VA2M} \\
        \cmidrule(lr){3-10}
        \multirow{4}{*}{\textbf{E}}
        & $\delta_m (g)$  & 253.28 & 0.00 & 0.00 & 0.00 & 273.14 & 0.00 & 0.00 & 0.00 \\
        & $\psi_f$   & 0.00 & 100.00 & 100.00 & 100.00 & 0.57 & 100.00 & 100.00 & 100.00 \\
        & $\bar{\psi}_f$   & 0.00 & 42.37 & 42.37 & 42.37 & 0.19 & 45.44 & 45.44 & 45.44 \\
        & $\Delta$  & 80.38 & 92.36 & 92.36 & 92.36 & 93.70 & 94.13 & 94.79 & 93.48 \\
        \midrule
        \multirow{4}{*}{\textbf{P5}}
        & $\delta_m (g)$  & \na & \na & \na & \na & 140.85 & -29.95 & -45.60 & 3.40 \\
        & $\psi_f$   & \na & \na & \na & \na & 13.16 & 64.97 & 51.87 & 95.22 \\
        & $\bar{\psi}_f$   & \na & \na & \na & \na & 27.10 & 1.79 & 5.55 & 0.25 \\
        & $\Delta$  & \na & \na & \na & \na & 0.00 & 92.79 & 92.17 & 0.00 \\
        \midrule
        \multirow{4}{*}{\textbf{P9}}
        & $\delta_m (g)$  & 118.96 & -36.53 & -63.10 & -99.53 & 109.07 & -51.17 & -91.35 & -0.31 \\
        & $\psi_f$   & 4.48 & 38.54 & 19.26 & 7.44 & 26.73 & 53.85 & 33.12 & 99.63 \\
        & $\bar{\psi}_f$   & 5.88 & 41.85 & 21.71 & 8.75 & NE & NE & NE & NE \\
        & $\Delta$  & 91.54 & 91.41 & 91.40 & 0.00 & 0.00 & 0.00 & 0.00 & 0.00 \\
        \midrule
        \multirow{4}{*}{\textbf{R5}}
        & $\delta_m (g)$  & 52.94 & 97.66 & -200.78 & -209.91 & \na & \na & \na & \na \\
        & $\psi_f$   & 35.65 & 14.92 & 2.00 & 1.68 & \na & \na & \na & \na \\
        & $\bar{\psi}_f$   & 15.56 & 9.71 & 1.99 & 1.38 & \na & \na & \na & \na \\
        & $\Delta$  & 93.60 & 93.62 & 0.00 & 0.00 & \na & \na & \na & \na \\
        \midrule
        \multirow{4}{*}{\textbf{R9}}
        & $\delta_m (g)$  & 103.68 & -154.11 & -99.85 & -403.24 & \na & \na & \na & \na \\
        & $\psi_f$   & 32.07 & 18.44 & 33.44 & 1.20 & \na & \na & \na & \na \\
        & $\bar{\psi}_f$   & 29.81 & 20.97 & 29.32 & 0.50 & \na & \na & \na & \na \\
        & $\Delta$  & 89.12 & 0.00 & 82.58 & 0.00 & \na & \na & \na & \na \\
        \midrule
        \multirow{4}{*}{\textbf{W5}}
        & $\delta_m (g)$  & -64.41 & -76.37 & -120.44 & -266.18 & -235.00 & -81.74 & 304.16 & -178.00 \\
        & $\psi_f$   & 35.44 & 29.23 & 14.38 & 1.38 & 8.91 & 43.13 & 4.38 & 16.02 \\
        & $\bar{\psi}_f$   & NE & NE & NE & NE & NE & NE & NE & NE \\
        & $\Delta$  & 0.00 & 0.00 & 0.00 & 0.00 & 0.00 & 90.59 & 97.68 & 94.90 \\
        \midrule
        \multirow{4}{*}{\textbf{W9}}
        & $\delta_m (g)$  & \na & \na & \na & \na & -289.49 & -210.37 & -118.32 & -324.61 \\
        & $\psi_f$   & \na & \na & \na & \na & 11.25 & 20.44 & 40.94 & 8.63 \\
        & $\bar{\psi}_f$   & \na & \na & \na & \na & 12.14 & 20.46 & 2.69 & 9.19 \\
        & $\Delta$  & \na & \na & \na & \na & 0.00 & 0.00 & 0.00 & 95.76 \\
        \specialrule{1.2pt}{0.2pt}{1pt}
    \end{tabular}
    \label{tab:forcesafecomp1}
    \vspace{-10pt}
\end{table}